\documentclass[letterpaper, 10 pt, conference]{ieeeconf}  

\IEEEoverridecommandlockouts                              

\usepackage[ruled,linesnumbered]{algorithm2e}
\usepackage{CJKutf8}
\usepackage{microtype}
\usepackage{longtable}
\usepackage{longtable}
\usepackage{graphicx}
\usepackage{subfigure} 
\usepackage{caption}

\usepackage{array}
\usepackage{url}
\usepackage{soul}
\usepackage{booktabs}
\usepackage{threeparttable}
\usepackage{multirow}
\usepackage{titlesec}
\usepackage{flushend}
\usepackage{color}
 \usepackage[bookmarks=true, colorlinks, citecolor=blue, linkcolor=black]{hyperref}
\usepackage{footmisc}
\usepackage{amssymb}

\titleformat*{\section}{\normalsize\bfseries}
\def\tsc#1{\csdef{#1}{\textsc{\lowercase{#1}}\xspace}}
\tsc{WGM}
\tsc{QE}
\tsc{EP}
\tsc{PMS}
\tsc{BEC}
\tsc{DE}
\usepackage{amsmath}

\title{\LARGE \bf
Robot-GST: geometry-aware spatial-temporal robot policy representation \\ and evaluation}

\author{Anonymous Author}

\author{
\parbox{0.9\textwidth}{\centering
Sichao Liu$^{1,*}$~
Zekun Wang$^{1}$~
Lixuan Tang$^{2}$~
Yiming Li$^{3,2}$~
Xiaohan Wang$^{4}$~
Hanzhi Zhang$^{1}$\\
Daqiang Guo$^{5}$~
Peng Zhou$^{6}$~
Lihui Wang$^{7,1}$
\thanks{$^{*}$Corresponding authors: Sichao Liu (sicliu@kth.se).}%
\\[0.5em]
{\small
$^{1}$KTH~
$^{2}$EPFL~
$^{3}$Idiap Research Institute~
$^{4}$Beihang University\\
$^{5}$The Hong Kong University of Science and Technology (Guangzhou)~
$^{6}$Great Bay University
$^{7}$The Hong Kong Polytechnic University
}%
}%
}

\begin{document}

\IEEEaftertitletext{%
\begin{center}
\includegraphics[width=\textwidth]{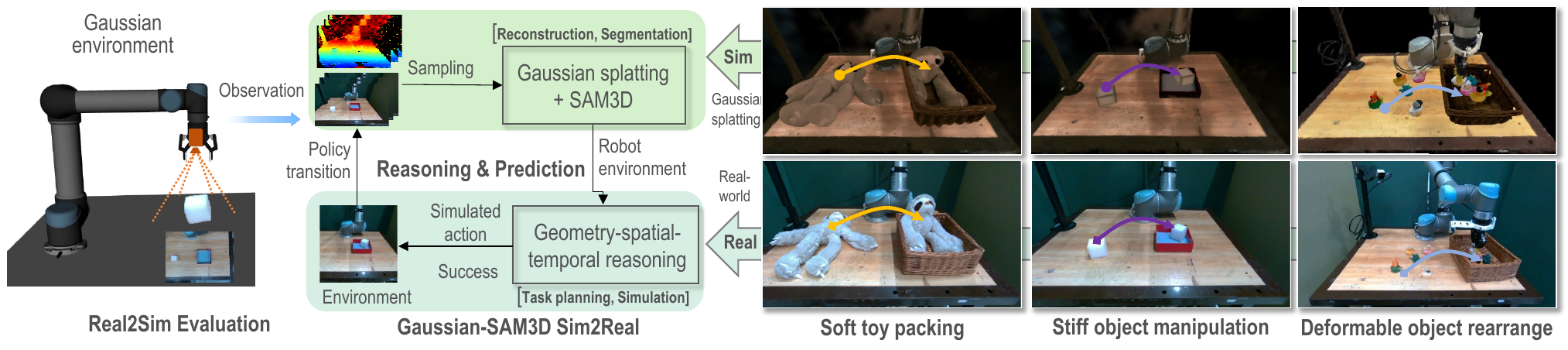}
\captionof{figure}{\textbf{Overview of Robot-GST: Geometry-aware Spatio-Temporal Policy Representation and Evaluation}.
Our system builds a unified, geometry-aware manipulation framework by combining 3D Gaussian Splatting with SAM3D segmentation, enabling prediction, reconstruction, part-aware reasoning, and robust sim-to-real transfer.}
\label{overview}
\end{center}
\vspace{-0.1em}
}

\maketitle
\thispagestyle{empty}
\pagestyle{empty}

\begin{abstract}
Robotic manipulation policies are advancing rapidly with increasing reliance on vision-language models for end-to-end decision making. However, reliable deployment remains challenging because many policies lack explicit mechanisms for predicting task outcomes and evaluating whether generated actions will achieve desired final states, causing execution errors to accumulate during long-horizon manipulation. We present Robot-GST, a geometry-aware spatio-temporal behaviour representation and evaluation framework that constructs a Gaussian-SAM robotic environment for real-to-sim policy verification and improves the reliability of real-world manipulation deployment. Our approach constructs a high-fidelity robotic environment from RGB-D observations using 3D Gaussian Splatting and SAM3D, enabling ``simulation and evaluation before acting''. It integrates visual observations and language instructions with spatio-temporal reasoning for long-horizon task planning using large vision-language models. To bridge high-level planning and real-world execution, we introduce Gaussian-aware final-state estimation through geometric sampling and state-based trajectory planning. Before execution, candidate action sequences are simulated and evaluated in the Gaussian-SAM environment to filter infeasible behaviours. We validate our approach on representative manipulation tasks involving rigid, soft, and deformable objects, including cube placing, toy packing, and duck rearrangement, demonstrating that geometry-aware spatio-temporal reasoning and state-aware execution improve manipulation reliability across different object categories. Our results suggest that combining geometry-aware reconstruction with high-quality rendering and simulation provides a scalable approach for evaluating robotic manipulation behaviours. Website: \url{https://robot-gst.github.io}
\end{abstract}

\section{Introduction}
Robot manipulation policies have advanced rapidly across a wide range of tasks, driven by recent progress in vision-based and vision-language-based learning approaches~\cite{chi2025diffusion, team2024octo, wu2026cei}. However, training large-scale robot policies remains expensive and challenging~\cite{kim2024openvla, intelligence2025pi}, despite their ability to directly map language and visual observations to robot actions. However, many existing policies lack explicit mechanisms for predicting post-action states, making it difficult to evaluate whether generated actions will achieve desired outcomes before execution~\cite{zhao2026difficulty}. Moreover, evaluating robot policies still largely relies on real-world experiments, which are time-consuming, costly, and difficult to scale~\cite{zhang2025real}. 

Vision-language-conditioned keypoint-affordance representations enable robotic manipulation by using pretrained models to generate semantic keypoints from visual observations and language instructions for path planning~\cite{huang2024rekep, liu2024moka, yuan2024robopoint}. However, transferring these generated paths to real-world environments remains challenging due to limited geometric awareness and sim-to-real discrepancies. These methods typically rely on sparse keypoints and stable correspondence, which can limit robustness under deformation, occlusion, or complex interactions~\cite{aljalbout2025reality}. In addition, many vision-language-guided task plans do not explicitly reason about object final states, limiting their application to tasks requiring collision-aware planning or contact-rich manipulation of deformable objects~\cite{din2025vision}.

Simulation before action provides a scalable approach that evaluates candidate behaviours before deployment on physical robots. \textit{Gaussian Splatting}~\cite{kerbl2023threedgs} represents scenes as explicit, anisotropic 3D Gaussians that can be rendered efficiently and support rapid editing and semantic annotation. Building on this representation, systems such as GraspSplats~\cite{huang2024graspsplats} fuse hierarchical, language-aligned features with depth supervision to enable open-vocabulary part localisation and real-time grasp proposals. More recently, real-to-sim evaluation pipelines~\cite{jiang2025phystwin} have combined photorealistic 3D Gaussian reconstructions with soft-body digital twins to correlate simulated rollouts with real-world success, providing scalable benchmarking but without coupling perception and reasoning for policy synthesis. Furthermore, physics-based simulators~\cite{jiang2025phystwin, abou2025real, badithela2025reliable} provide a framework for reconstructing deformable objects, simulating interactions, and evaluating policies.

We propose a geometry-aware spatio-temporal behaviour representation and evaluation framework that integrates visual observations and language instructions with spatio-temporal reasoning for state-aware task planning. After generating keypoint-based manipulation trajectories, we construct a Gaussian-SAM3D robotic simulation environment to evaluate candidate behaviours before execution. We evaluate our method on representative manipulation tasks, including rigid cube placement, soft-toy packing, and deformable-object rearrangement. Compared with existing methods, our approach significantly improves the alignment between simulated and real-world environments, achieving a structural similarity index measure (SSIM) of 0.044. Furthermore, evaluating policies in the photorealistic simulation environment improves the average real-world task success rate by 44.67\% after policy correction, confirming the practical benefits of our approach for sim-to-real robot policy evaluation. 

Our contribution can be summarised as follows:
\begin{itemize}
\item \textbf{A Gaussian-SAM3D-based simulation and evaluation environment}, built from RGB-D images using 3D Gaussian Splatting and SAM3D, enabling high-fidelity real-to-sim evaluation with photorealistically reconstructed objects;
\item \textbf{A geometry-aware spatio-temporal behaviour representation for multi-stage manipulation}, which embeds visual observations and language tasks into spatio-temporal reasoning to produce high-level plans that explicitly account for object geometry and future state evolution;
\item \textbf{Gaussian-aware final-state estimation and state-based low-level trajectory planning}, in which candidate action sequences are simulated and evaluated in the simulation environment before execution, enabling near-perfect real-world deployment across tasks.
\end{itemize}

\section{Related Work}
\label{sec:relatedwork}
\subsection{Language task decomposition and relational reasoning}
Large Language Models (LLMs) provide strong semantic priors for decomposing natural-language instructions into executable subgoals and reasoning about task dependencies~\cite{ahn2022saycan,pmlr-v205-huang23c,liang2023code}.
Beyond explicit planning, recent vision–language–action (VLA) policies leverage large-scale pretraining to execute diverse real-world tasks from language and visual inputs while reducing dependence on task-specific supervision~\cite{brohan2023rt}. Frameworks such as ReKep~\cite{huang2024rekep} formalise long-horizon manipulation as relational keypoint constraints, automatically generating constraint code and keypoints from language and RGB-D inputs. However, these methods primarily reason over sparse geometric structures and do not explicitly maintain dense 3D object representations that capture object parts, shapes, and evolving physical states. Mark-based visual prompting (MOKA) enables open-world planning with minimal training by converting image-space marks into structured waypoints~\cite{liu2024moka}. SoFar~\cite{qi2025sofar} further connects language instructions with 6-DoF semantic orientations, such as insertion directions, to enforce geometric alignment constraints. However, such approaches generally do not explicitly model object final states, limiting their application to tasks requiring collision-aware planning or contact-rich manipulation.

\subsection{Robotic scene representation and reconstruction}
Recent advances in 3D scene modelling have transformed robotic perception. Implicit radiance-field methods~\cite{mildenhall2022nerf,lerf2023} unify appearance and semantics across views but remain computationally expensive and difficult to edit, limiting their use for real-time manipulation. In contrast, \textit{3D Gaussian Splatting (3DGS)}~\cite{kerbl2023threedgs, Wu_2024_CVPR} represents scenes as explicit, anisotropic Gaussians that can be updated or rendered at interactive rates. Recent feature-augmented variants extend 3DGS by distilling dense foundation-model features for downstream semantics~\cite{zhou2023feature3dgs}. Extensions such as GraspSplats~\cite{huang2024graspsplats} fuse hierarchical, language-aligned features with depth supervision to achieve open-vocabulary part localisation and real-time grasp sampling. These explicit models offer high efficiency and editability for dynamic scenes, while SLAM-oriented 3DGS systems further enable continuous scene updates for robotics~\cite{huang2024gs_slam,li20254d}. Nevertheless, these representations remain largely perception-centric and provide limited mechanisms for structured temporal or relational reasoning required for multi-step interaction planning.

\subsection{Policy simulation and evaluation }
Recent 3DGS-based pipelines reconstruct scenes from real-world captures and couple them with approximate dynamics to estimate sim-to-real correlation and to select or refine candidate behaviours before deployment~\cite{jiang2025phystwin,qureshi2024splatsim}. Beyond individual pipelines, several recent works have begun systematising real-to-sim evaluation for robot manipulation policies through standardised benchmarks and reusable environment-generation tools~\cite{sedlacek2025realm,fang2025rebot}. These approaches exploit 3DGS as a efficient renderer for policy observations while relying on auxiliary geometric proxies or physics engines to simulate contact-rich interactions. Digital twins have also proven valuable for reconstructing and simulating real-world objects~\cite{abou2025real}. However, existing frameworks often struggle to capture complex real-world dynamics because they rely on predefined physical parameters or approximate interaction models~\cite{zhang2025real}. Physics-informed simulators, therefore, provide a promising alternative. For example, PhysTwin builds physical digital twins from sparse video inputs, enabling accurate reconstruction and resimulation of deformable objects~\cite{jiang2025phystwin}. However, existing approaches typically focus on individual components, such as semantic scene reconstruction, task reasoning, or simulation, without integrating geometry-aware representation, long-horizon reasoning, and pre-execution behaviour evaluation into a unified framework.

\section{Methods}
We describe how Gaussian representations and SAM3D construct a geometry-aware robotic environment for manipulation planning and evaluation. Our approach addresses two challenges: (i) obtaining high-fidelity geometric and semantic scene representations, which we address via 3DGS reconstruction and multimodal feature alignment, and (ii) evaluating rigid and deformable object manipulation behaviours within the reconstructed GS environment.  Addressing the second gap requires ensuring collision-free motion between rigid objects and the environment (including contacted objects), while deformable-object (including contacted objects), while deformable-object manipulation must satisfy constraints imposed by final-state estimation.

\subsection{3D Gaussian Splatting for robotic scene reconstruction}
\label{sec:3dgs_robotic_scene}

\textbf{Gaussian Splatting Reconstruction}: To reconstruct a photorealistic and geometry-consistent robotic scene from short multi-view RGB-D scans, we adopt 3D Gaussian Splatting (3DGS) as the scene representation. We model the scene as a collection of anisotropic Gaussian primitives:
\begin{equation}
\mathcal{G}=\{g_i\}_{i=1}^{N}, \qquad
g_i = (\boldsymbol{\mu}_i,\,\boldsymbol{r}_i,\,\boldsymbol{\Sigma}_i,\,\boldsymbol{c}_i,\,\alpha_i,\,\mathbf{f}_i)
\end{equation}
where $\mathcal{G}=\{g_i\}_{i=1}^{N}$ denotes a set of $N$ Gaussians. $\boldsymbol{\mu}_i\!\in\!\mathbb{R}^3$ denotes the Gaussian centre, $\boldsymbol{r}_i\!\in\!\mathbf{SO}(3)$ encodes its orientation, $\boldsymbol{\Sigma}_i\!\in\!\mathbb{R}^{3\times3}$ denotes its covariance, and $(\mathbf{c}_i\!\in\!\mathbb{R}^3,\,\alpha_i\!\in\![0,1])$ denote the colour and opacity, respectively.

We further associate each Gaussian with a semantic feature vector $\mathbf{f}_i\!\in\!\mathbb{R}^{d_f}$ extracted from foundation vision models (e.g., SAM3D, DINOv2, and CLIP). Given a query $y$, object- and part-level activations $s_i^{\text{obj}}(y)$ and $s_i^{\text{part}}(y)$ are derived from $\mathbf{f}_i$ for language-conditioned segmentation and manipulation.

Given calibrated cameras and a set of images $\{I_i\}_{i=1}^{N}$, we render the scene with a differentiable splatting renderer using front-to-back alpha compositing:

\begin{equation}
\begin{aligned}
I_i^* = \sum_i \boldsymbol{T}_i(u)\,\alpha_i\,\boldsymbol{w}_i(u)\,\boldsymbol{c}_i,  \\ \boldsymbol{T}_i(u) = \sum_{j<i}\bigl(1 - a_j(u)\bigr), a_i(u) = \alpha_i\,\boldsymbol{w}_i(u)
\end{aligned}
\label{eq:alpha_comp}
\end{equation}
where $\boldsymbol{w}_i(u)$ is the 2D splat weight at pixel $u$ (obtained by projecting the 3D Gaussian), and $\boldsymbol{T}_i(u)$ is the accumulated transmittance along the viewing ray.
We optimise the Gaussian parameters using a reconstruction objective $\mathcal{L}$, optionally regularised with a depth prior and feature distillation, weighted by $\lambda_d$ and $\lambda_f$.


\begin{equation}
\mathcal{L}=\sum_{i=1}^{N} \|I_i^*-I_i\|_2^2
+\lambda_d\|D-D^*\|_1
+\lambda_f\|F-F^*\|_2^2
\label{eq:3dgs_loss}
\end{equation}
where $I^*$ is the rendered image, $D^*$ is the rendered depth, $D$ is an aligned dense depth prior that stabilises geometry under sparse views, and $F$ and $F^*$ denote rendered and target semantic features for object-centric interaction.

\textbf{Mesh model}: Given an input image $I$, we obtain an object mask $M$ using SAM3. We then follow SAM3D to construct a geometric model $p(O, R, t, s\mid I, M)$, where $O \in \mathbb{R}^{64^3}$ denotes a coarse (voxelised) object shape, $R \in \mathbb{R}^{6}$ is a 6D rotation representation, $t \in \mathbb{R}^{3}$ is the translation, and $s \in \mathbb{R}^{3}$ is the scale. Importantly, SAM3D incorporates calibrated depth observations, enforcing metric scale during object reconstruction and ensuring that the estimated translation $t$ and scale $s$ match real-world object dimensions. Active voxels from the coarse shape $O$ are used to refine the object shape $S$ and texture $T$ by learning the conditional distribution $p(S, T\mid I, M, O)$. The latent representation $L$ of $S$ and $T$ is then decoded into a mesh $\mathcal{M}$ via a mesh decoder $D_m$ and into Gaussians $\mathcal{G}$ via a Gaussian-splat decoder $D_g$:
\begin{equation}
\mathcal{M}=D_m(L(S,T)); \; \mathcal{G}=D_g(L(S,T))
\label{eq:mesh_and_gs_decode}
\end{equation}
The mesh $\mathcal{M}$ provides explicit geometry for collision reasoning, while $\mathcal{G}$ enables efficient photorealistic rendering and scene composition. Compared with meshes produced by SAM3D alone, the resulting mesh offers a favourable trade-off between fidelity, accuracy, and reconstruction time. The static GS environment, mesh-based objects, and the robot model enable compositional scene construction, allowing the robot to manipulate objects in a dynamic environment.

\subsection{Keypoint affordance constraints}
\label{sec:vlm_rekep}
\textbf{Keypoint and relational constraints}: Given a natural-language task $l$, we process annotated scene images $I$ (with object masks $M$ and labels) using a vision-language model to generate a keypoint-affordance representation $\mathbf{k}$. We construct a discrete set of scene keypoints:
\begin{equation}
\mathbf{k}= \mathcal{F}(I,M,l,D); \; \mathbf{k} = [\mathbf{k}_1,\dots,\mathbf{k}_K] \in \mathbb{R}^{K\times 3}
\end{equation}
where each $\mathbf{k}_i$ denotes a 3D point associated with a target object $O$. We express a relational constraint as a scalar-valued function:
\begin{equation}
f:\mathbb{R}^{K\times 3}\rightarrow\mathbb{R}, \qquad f(\mathbf{k}) \le 0 \;\Rightarrow\; \text{constraint satisfied}.
\end{equation}
Here, $f$ encodes proximity (distance-to-contact), alignment (axis/normal consistency), or relative-pose relations between keypoints (e.g., ``rearrange deformable toys from the table into a basket''). Each stage is represented by two types of constraints: (i) \emph{sub-goal} constraints $\mathcal{C}^{(i)}_{\mathrm{sub}}=\{f^{(i)}_{\mathrm{sub},1},\dots,f^{(i)}_{\mathrm{sub},n_i}\}$, which specify grasp and release conditions, and (ii) \emph{sub-path} constraints $\mathcal{C}^{(i)}_{\mathrm{path}}=\{f^{(i)}_{\mathrm{path},1},\dots,f^{(i)}_{\mathrm{path},m_i}\}$, which impose \emph{path-feasibility} requirements during motion. Importantly, the VLM generates executable constraint representations, avoiding task-specific hand-crafted logic and enabling downstream optimisation.

\textbf{Spatial constraints with task decomposition}: We decompose a language task into a sequence of stages $\mathcal{H}=\{1,\dots,H\}$, each represented by relational keypoints and geometric constraints. For stage $h$, let $\mathcal{K}_h=\{\mathbf{k}_m\}\subset\mathbb{R}^3$ denote the task-relevant keypoints for the target object $O$, and let $\mathcal{C}_h=\{c_\ell\}$ denote a set of differentiable spatial constraints encoding subgoals such as distance, alignment, orientation, or visibility. Each stage defines an optimisation problem over the decision variables $\boldsymbol{\zeta}_h=(T_h,\gamma_h)$:

\begin{equation}
\label{eq:stage_optim}
\begin{aligned}
\zeta_h^{\star}
&= \arg\min_{\zeta_h} \; J_h(\zeta_h \mid \mathcal{G}, l) \\
&= \sum_{\ell=1}^{L_h} w_{h,\ell}\;
\phi_{h,\ell}\!\Big(
\big[f_{h,\ell}(K_h,\zeta_h;\mathcal{G},l)\big]_+
\Big)
\;+\;
\lambda_{\mathrm{reg}}\;\Omega(\zeta_h)
\end{aligned}
\end{equation}
where $T_h \in \mathrm{SE}(3)$ and $\gamma_h$ denote the end-effector pose (position and orientation) and grasp/gripper parameters at stage $h$. Here, $L_h$ is the number of constraint terms. The $\ell$-th inequality constraint function $f_{h,\ell}$ is satisfied when $f_{h,\ell}(\cdot)\le 0$. $\phi_{h,\ell}(\cdot)$ is a nonnegative penalty on constraint violation, and $w_{h,\ell} > 0$ is the corresponding weight. $\Omega(\zeta_h)$ is a regulariser on $\zeta_h$ (e.g., smoothness or deviation from a nominal pose/grasp), with weight $\lambda_{\mathrm{reg}}$. Finally, $[x]_+ = \max(0,x)$ denotes the violation magnitude of an inequality constraint (zero if satisfied).

\subsection{Geometry-aware spatio-temporal reasoning and planning}
\label{sec:geometry-aware}
\textbf{Object state prediction}: We represent object $O_i$ by its world-frame translation $\mathbf{p}_i\in\mathbb{R}^3$ and unit-quaternion orientation $\mathbf{q}_i\in\mathbb{H}$ (with $\mathbb{H}$ denoting the set of unit quaternions). The relational-spatial constraints define target grasp and release positions $(\mathbf{p}_{\mathrm{grasp}},\mathbf{p}_{\mathrm{release}})$. Let $\mathbf{p_c}_i\in\mathbb{R}^3$ denote a representative object centre (e.g., its centroid) used for association. We select the target object index $t$ via nearest-neighbour matching:
\begin{equation}
O_t = \arg\min_{O_i}\|\mathbf{p_c}_i-\mathbf{p}_{\text{grasp}}\|_2.
\end{equation}
To approximate stable object transport, we assume the object–end-effector relative transform remains fixed after grasping. Let $(\mathbf{p}_{ee},\mathbf{q}_{ee})$ be the current end-effector pose, and let $(\mathbf{p}_0,\mathbf{q}_0)$ be the grasped object's pose at the grasp moment. Let $R_{ee}\in\mathbb{R}^{3\times 3}$ be the rotation matrix corresponding to $\mathbf{q}_{ee}$, and let $\otimes$ denote quaternion multiplication. Then:
\begin{equation}
\mathbf{p}_{\text{rel}} = R^{-1}_{ee}(\mathbf{p}_0-\mathbf{p}_{ee}), \quad
\mathbf{q}_{\text{rel}} = \mathbf{q}_{ee}^{-1}\!\otimes\!\mathbf{q}_0.
\end{equation}

To make placement robust to reconstruction noise, partial occlusion, and narrow receptacles, we perform a local search around a nominal end-effector pose predicted by the stage constraints.
Let $\mathbf{p}_{\mathrm{final}}\in\mathbb{R}^3$ and $\mathbf{q}_{\mathrm{final}}\in\mathbb{H}$ denote the nominal end-effector translation and orientation. We generate candidate end-effector poses by sampling bounded positional perturbations $\Delta\mathbf{p}$:
\begin{equation}
\mathbf{p}_{\mathrm{rand}} = \mathbf{p}_{\mathrm{final}} + \Delta \mathbf{p},
\qquad
\Delta \mathbf{p} \sim \mathcal{U}(-\tau_p,\tau_p)^3,
\end{equation}
where $\mathcal{U}$ is the uniform distribution and $\tau_p>0$ bounds the translation noise per axis. We also sample a bounded rotational perturbation $\Delta\boldsymbol{\phi}\in\mathbb{R}^3$ in axis--angle form with $\|\Delta\boldsymbol{\phi}\|_2\le\tau_r$ (for a scalar bound $\tau_r>0$), and apply it to obtain a perturbed orientation $\mathbf{q}_{\mathrm{rand}}$. Each candidate thus defines a hypothesised release pose for the manipulated object.

\textbf{Candidate sampling and collision checking}: We explore feasible release poses by sampling bounded perturbations $\Delta\mathbf{p}$ and rotational noise $\Delta\boldsymbol{\phi}$. For each candidate, we evaluate geometric feasibility by checking for collisions with the static scene geometry. We build a signed distance field (SDF) $\mathcal{S}:\mathbb{R}^3\to\mathbb{R}$, where $\mathcal{S}(\mathbf{x})$ denotes the distance from point $\mathbf{x}$ to the closest scene surface (positive in free space). For a set of object surface points $\{\mathbf{x}_m\}$ transformed by the candidate pose (as implied by the end-effector pose and the fixed relative transform), we compute
\begin{equation}
d(\mathbf{x}_m)=\mathcal{S}(\mathbf{x}_m),
\end{equation}
and declare a candidate collision-free if
\begin{equation}
\min_{\mathbf{x}_m} d(\mathbf{x}_m) \ge \epsilon_{\mathrm{col}},
\end{equation}
where $\epsilon_{\mathrm{col}}>0$ is a collision margin.
Among feasible candidates, we rank solutions using a stability-oriented heuristic (e.g., minimal rotational deviation from a desired target orientation) and then invoke the motion planner to generate a collision-free trajectory that satisfies both the stage subgoal constraints and the path constraints.

\subsection{Robot behaviour simulation and evaluation}
We verify candidate low-level action sequences in a Real-to-Sim environment that is \emph{metrically aligned} with the real scene (i.e., the same scale and robot base frame), enabling direct trajectory verification without additional normalisation.
The environment maintains complementary representations: (i) a Gaussian scene for photorealistic rendering and observation consistency, and (ii) an explicit geometric proxy (mesh/TSDF/SDF extracted from the reconstruction and SAM3D objects) for collision and contact reasoning.

\textbf{Trajectory parameterisation.} We represent a candidate manipulation trajectory as
\begin{equation}
\tau = \{u_t\}_{t=1}^{T}, \qquad u_t = (\mathbf{p}_t,\mathbf{q}_t,g_t),
\end{equation}
where $\mathbf{p}_t\in\mathbb{R}^3$ and $\mathbf{q}_t\in\mathbb{H}$ denote the commanded end-effector position and unit-quaternion orientation in the robot base frame, and $g_t\in\{0,1\}$ is the gripper open/close command.

\textbf{Real-to-Sim rollout model}: Let $x_t$ denote the simulator state (robot joint configuration, gripper state, and object states) at step $t$.
We compute joint targets via inverse kinematics
\begin{equation}
\mathbf{\theta}_t = \mathrm{IK}(\mathbf{p}_t,\mathbf{q}_t),
\end{equation}
then advance the state using a simulator transition model
\begin{equation}
 x_{t+1} = F(x_t,\mathbf{\theta}_t,g_t),
\end{equation}
where $F$ combines URDF kinematics with physics/contact simulation.
Rigid objects are simulated with rigid-body dynamics and collision meshes reconstructed from the scene/object geometry; deformable objects are simulated with a soft-body model (PhysTwin~\cite{jiang2025phystwin}) to approximate deformation and contact response.

\textbf{Verification criteria (simulate before acting):}
We consider $\tau$ feasible if it is executable, collision-safe, and satisfies the task goal at the terminal state.
We define the acceptance indicator
\begin{equation}
\mathbb{I}(\tau)=\prod_{t=1}^{T} \mathbb{I}\big(\mathrm{IK}(\mathbf{p}_t,\mathbf{q}_t)\ \big)
\cdot \prod_{t=1}^{T} \mathbb{I}\big(\delta(x_t)\ge \epsilon_{\mathrm{col}}\big)
\cdot \mathbb{I}\big(\Psi(x_T,l)=1\big)
\end{equation}
where $\delta(x_t)$ is the minimum signed distance between all rigid bodies (robot, gripper, rigid objects) and the static scene geometry at step $t$, computed from the reconstructed mesh/TSDF/SDF proxy; $\epsilon_{\mathrm{col}}>0$ is a safety margin; and $\Psi(x_T,l)$ is a task-specific terminal predicate evaluated on the final simulated state.
We execute feasible trajectories $\tau$ on the real robot when $\mathbb{I}(\tau)=1$.

\section{Experiments}

\textbf{Experimental setup}:
The robotic setup consists of a UR5 robot arm equipped with wrist and side RealSense cameras (D435 and D405), connected to a workstation with an NVIDIA GeForce RTX 4090 GPU. We evaluate our method on three manipulation tasks involving rigid, soft, and deformable objects:
\begin{itemize}
\item 1) Cube placing: the robot grasps a rigid cube from the table and places it into a small box. A trial is considered successful if the cube is placed inside the box without collision.
\item 2) Toy packing: the robot picks up a soft toy (sloth) from the table and places it into a basket. Success is defined as the toy being fully placed in the basket.
\item 3) Deformable toy rearrangement: the robot picks up a set of deformable duck toys from the table and rearranges them in a basket. Placing all toys into the basket is considered a success.
\end{itemize}


\begin{figure}[t]
\vspace{0.5em}
    \centering
    \includegraphics[width=0.5\textwidth]{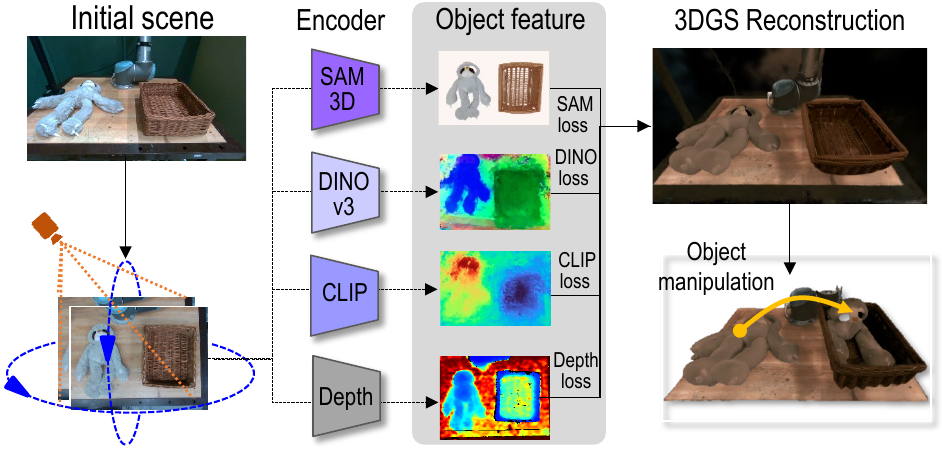}
    \caption{\textbf{3DGS reconstruction of robotic scenes.}
    Multi-view RGB-D observations are processed using four complementary encoders: SAM3D for object-level segmentation, DINOv3 for semantic correspondence, CLIP for vision--language alignment, and depth cues for geometric structure. The resulting features enable high-fidelity 3D scene reconstruction.
}
\label{fig:3dgs}
\vspace{-1.5em}
\end{figure}

To create a robotic environment for behaviour verification and evaluation, we adopt a compositional scene-construction approach. We first construct the static environment from visual and geometric data, then build the robot's virtual representation, and finally create mesh-based representations of the manipulated objects.

\textbf{3DGS reconstruction}: As shown in Fig.~\ref{fig:3dgs}, the scene is captured from multiple viewpoints to obtain RGB images and depth measurements, enabling consistent modelling of appearance and metric geometry. We obtain object-level segmentation using SAM3D by lifting and fusing 2D masks into a coherent 3D representation. DINOv3 features provide semantic correspondence across views, while CLIP introduces language-aligned semantic information. Depth cues impose explicit geometric constraints. We integrate these multimodal features into an object-centric representation for geometry-aware scene reconstruction. We use the resulting representation to reconstruct the scene via 3D Gaussian Splatting, producing a continuous, photorealistic 3D model. Since all components are represented as Gaussian primitives, we align the static scene, the robot, and the objects within a shared world frame.

We align the reconstructed GS scene with the robot URDF model using ICP, enabling consistent pose mapping between real and simulated environments. We sample points per link to ensure sufficient coverage. Before ICP, we use SuperSplat to adjust the scan pose such that the centre of the robot base frame coincides with the origin of the world frame. The static background GS is in the same frame as the robot GS, and we apply the same ICP-estimated transform. After aligning the static background, the robot, and the objects into a unified GS scene, forward kinematics updates the robot state in both joint and Cartesian spaces. During task execution (e.g., packing a sloth), the pose of the real robot is mapped to its counterpart in the GS scene: joint and gripper meshes are updated according to the robot’s rigid-body motion, and the robot in the GS environment is driven by the command sequence produced by the robot policy to evaluate task performance.

\textbf{Metrics}: To evaluate GS reconstruction quality, we report standard photometric rendering metrics (PSNR, SSIM, and LPIPS) in Table~\ref{rendering-comparison}~\cite{sandstrom2023point}. The best result is in bold, and the second-best is underlined.

\textbf{Baseline methods}: We compare our reconstruction with representative 3DGS-based approaches. Specifically, we compare against GScream~\cite{wang2024learning}, VR-GS~\cite{jiang2024vr}, Decoupled GS~\cite{wang2025decoupledgaussian}, POGS~\cite{yu2025persistent}, D3DGS~\cite{luiten2024dynamic}, and Embodied GS~\cite{abou2024physically} on a robotic scene. We select these methods based on publicly reported results for robotic scene reconstruction or on the availability of source code for benchmarking.

\vspace{-4pt}
\begin{table}[h]\centering
    \caption{\textbf{Average rendering performance on robotic scene reconstruction}. Our method achieves competitive performance across rendering metrics. } \label{rendering-comparison}    
        \begin{tabular}{@{} >{\raggedright\arraybackslash}p{0.15\textwidth}>
        {\raggedright\arraybackslash}p{0.08\textwidth}>
        {\raggedright\arraybackslash}p{0.08\textwidth}>
        {\raggedright\arraybackslash}p{0.08\textwidth}
        @{} }     
            \toprule
            \textbf{Methods}   & \textbf{PSNR[db]$\uparrow$} & \textbf{SSIM$\uparrow$} & \textbf{LPIPS(vgg)$\downarrow$} \\ 
            \midrule
              GScream~\cite{wang2024learning} & 17.82 & 0.590 & 0.56 \\
              VR-GS~\cite{jiang2024vr} & 24.13 & 0.833 & 0.32 \\
              Decoupled GS~\cite{wang2025decoupledgaussian} & \textbf{27.32}& \underline{0.905} & 0.30 \\
              POGS ~\cite{yu2025persistent} & 19.66 & 0.869 & \underline{0.098} \\
              D3DGS~\cite{luiten2024dynamic} & 12.05 & 0.399 & 0.23 \\
              Embodied GS~\cite{abou2024physically} & 14.95 & 0.389 & 0.36 \\
     \textbf{Ours} & \underline{24.64} & \textbf{0.963} & \textbf{0.044}  \\
            \bottomrule
        \end{tabular}
        \vspace{-4pt}
    \end{table}

\begin{figure}[h]
    \centering
    \includegraphics[width=0.5\textwidth]{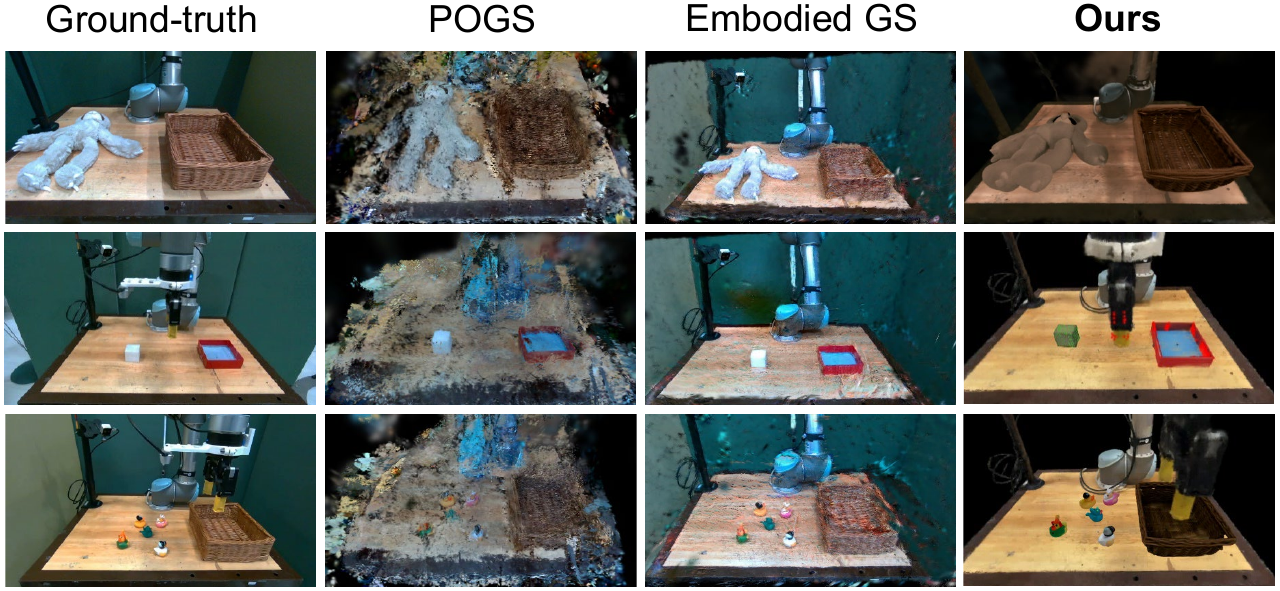}
    \caption{\textbf{GS reconstruction rendering examples}: Our method achieves the best GS reconstruction performance compared with POGS and Embodied GS.
    } 
    \label{gs-reconstruction}
\end{figure}

Our method, with an SSIM of 0.963 and an LPIPS of 0.044, outperforms the baselines and ranks second on PSNR (24.64), offering a significant advantage for high-fidelity reconstruction of the robotic environment. Fig.~\ref{gs-reconstruction} provides a qualitative comparison of our approach with POGS and Embodied GS (both of which use robotic tasks for benchmarking). Our method produces high-quality GS reconstructions with reduced artefacts in objects and robotic scenes, which are critical for policy simulation and evaluation. In contrast, POGS and Embodied GS struggle to render fine details and exhibit substantial noise in the reconstructed environment.

\begin{figure}[h]
    \centering
    \includegraphics[width=0.5\textwidth]{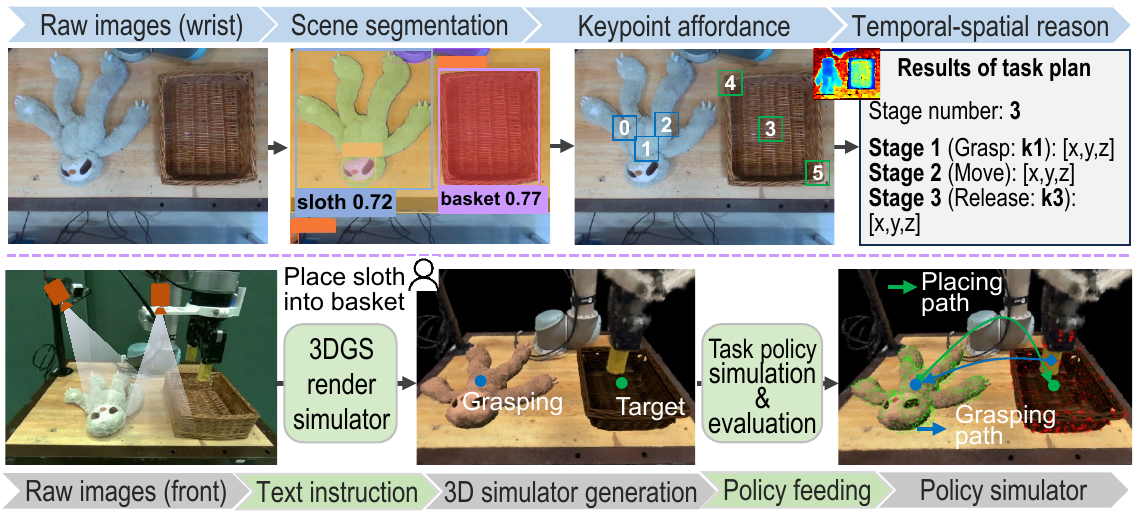}
    \caption{\textbf{Geometry-aware spatio-temporal planning and policy evaluation}. Top: vision--language-conditioned keypoint-affordance representation and task planning in a 3D coordinate frame; bottom: policy evaluation in the constructed simulator using reconstructed robotic scenes.
    } 
    \label{task-plan}
    \vspace{-1em}
\end{figure}

\textbf{Geometry-aware spatio-temporal reasoning and planning}: 
As shown in Fig.~\ref{task-plan}, given a natural-language task (e.g., packing a sloth into a basket), the system feeds an image captured by the wrist camera into a segmentation network. The system segments, masks, and labels task-relevant objects (e.g., the sloth and basket) with text and bounding boxes to produce an annotated image. The annotated image is then fed into a large language model (GPT-5) to generate keypoint candidates for object manipulation based on object geometry, with grasping and target keypoints indicated by points 1, 2, 3, 4, 5, and 6, respectively. The pre-trained model is then prompted to decompose the language task into $\mathcal{H}=3$ subtasks (stages), each described in text. Here, we follow the prompt-design practice of~\cite{liu2024moka}. For each subtask, we use 2D keypoint affordances, depth observations, and pose relations as inputs to subgoal and subpath constraint functions. Our method estimates feasible final object states by sampling end-effector poses while holding it, and candidates are sampled as described in Sec.~\ref{sec:geometry-aware}. These candidates chain keypoint-based subgoals together to generate robot trajectories.

Before execution on the real robot, we evaluate the generated manipulation trajectory in a 3DGS-based simulator, as shown at the bottom of Fig.~\ref{task-plan}. The robot follows the planned trajectory (blue line) to grasp the object (e.g., a sloth) at the grasp keypoint (blue point: keypoint 1). The gripper closes at the grasp point, after which the robot moves the object along the placement trajectory (green line). When part of the sloth (a soft object) contacts the target object (the basket), the embodied physics constraints of the PhysTwin-based simulator deform the contacted body part inside the basket. Specifically, the robot places the sloth's main body into the basket at the target point (green point: keypoint 3). Other body parts (e.g., legs and arms) contact the basket edge, and simulated contact forces deform them within the basket. If the simulation is successful, the policy can be deployed on the real robot.

\begin{figure}[h]
    \centering
    \includegraphics[width=0.5\textwidth]{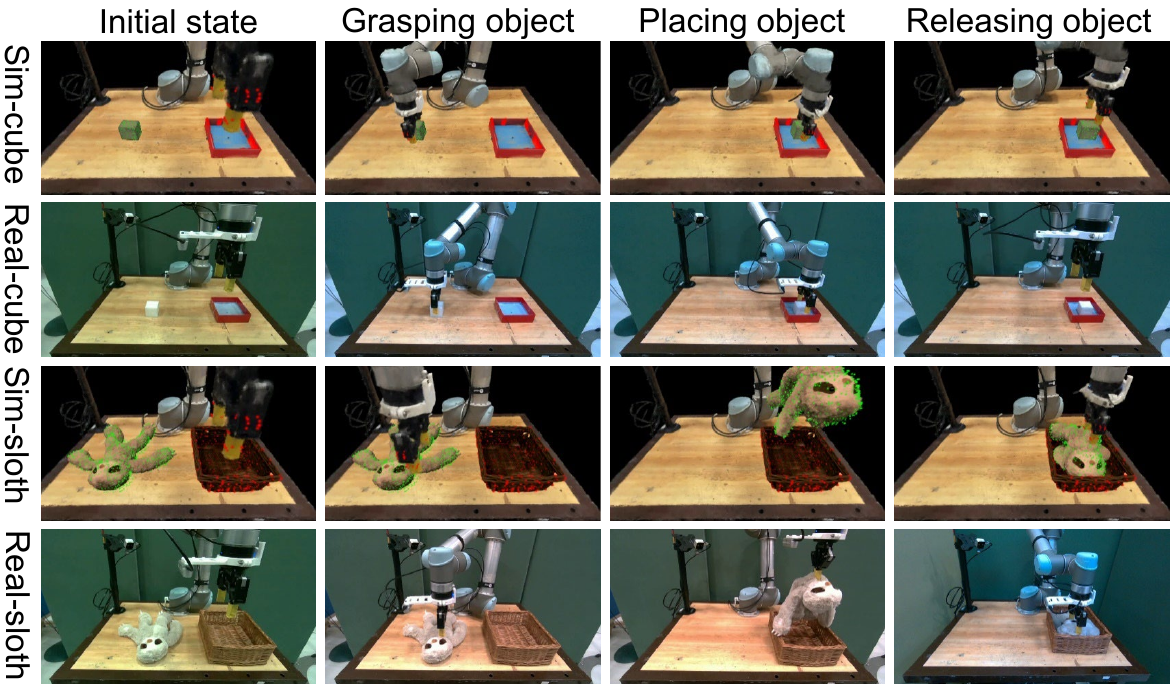}
    \caption{\textbf{Simulation and real-world experiments for rigid and soft-object manipulation}. Rows 1 and 3 show simulation results, while Rows 2 and 4 show real-world results for cube placing and sloth packing, respectively.
    } 
    \label{experiment-cube}
\end{figure}

\textbf{Robot policy evaluation in the simulator and rollout on the real robot}: Fig.~\ref{experiment-cube} shows results of behaviour evaluation in the simulator for cube packing and sloth packing (Rows 1 and 3). The point clouds of the cube (green) and box (red) are used for collision checking while the robot follows the generated trajectories. Specifically, our method measures the distances between the cube and box point clouds, with distances below a predefined threshold indicating a collision. The simulation results show that the robot can place the cube into the box by following the generated trajectory. The robot then executes the corresponding trajectory in the real world and successfully completes the task (Row 2). Similarly, we evaluate the sloth-packing policy in simulation before real-world rollout. Compared with rigid-object manipulation, deformable-object manipulation relies more heavily on contact interactions. The simulator therefore applies contact forces to deform the object and keep it within the basket. However, the simulated contact forces cannot exactly match those in the real world, which can lead to discrepancies between simulation and real-world deployment, particularly for body-part contact.

\begin{figure}[h]
    \centering
    \includegraphics[width=0.5\textwidth]{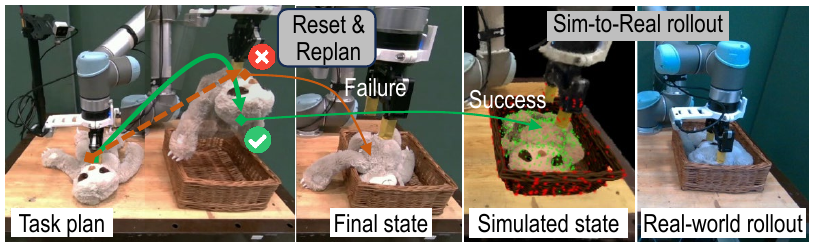}
    \caption{\textbf{Policy evaluation and replanning}. A robot policy (orange) fails (the sloth is not fully packed into the basket), triggering replanning (green), which the simulator evaluates. If successful, the policy is rolled out on the robot.
    } 
    \label{policy-replan}
 
\end{figure}

To address discrepancies between simulation and real-world rollout, we implement a behaviour evaluation and replanning scheme that enables the robot to recover from failures. As shown in Fig.~\ref{policy-replan}, a policy evaluated in simulation (orange) fails in the real world, leaving part of the sloth outside the basket. The control script then resets the scene and replans the task. The new plan (green) is evaluated in the simulator, and the results indicate that the sloth can be placed in the basket. Finally, the sim-to-real rollout on the real robot succeeds.

\begin{table}[h]
\centering
\caption{\textbf{Success rate of tasks with/without replanning in real and simulation environments.}}
\label{success}
\begin{tabular}{@{}lcccc@{}}
\toprule
 & \multicolumn{2}{c}{\textbf{Without replanning}} 
 & \multicolumn{2}{c}{\textbf{Replanning}} \\ 
\cmidrule(lr){2-3} \cmidrule(lr){4-5}
\textbf{Task} & \textbf{sim} & \textbf{real} & \textbf{sim} & \textbf{real} \\
\midrule
Place cube into a box      & 80.0\% & 80.0\% & 100.0\% & 90.0\% \\
Package sloth into a basket     & 60.0\% & 40.0\% & 100.0\% & 80.0\% \\
Arrange ducks into a basket     & 60.0\% & 46.0\% & 80.0\% & 70.0\% \\

\midrule
\textbf{Average}  & \textbf{66.7\%} & \textbf{55.3\%} 
               & \textbf{93.3\%} & \textbf{80.0\%} \\
\bottomrule
\end{tabular}
\end{table}

\begin{figure}[h]
    \centering
    \includegraphics[width=0.5\textwidth]{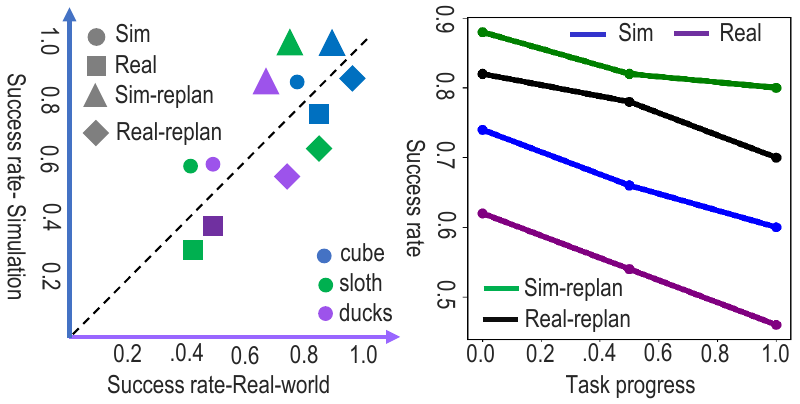}
    \caption{Left: \textbf{Correlation of success rate in simulation and real-world experiments}; Right: \textbf{Success rate of each control step (subtask)} measured on the physical robot over 10 trials for toy duck rearrangement.
    } 
    \label{success-rate}
    \vspace{-1em}
\end{figure}

For each task, we record success over 10 trials, varying object positions across trials. If any subtask in the sequence fails (e.g., duck rearranging), subsequent subtasks are not executed. Fig.~\ref{success-rate} (left) shows the correlation between success rates in simulation and real-world experiments across all tasks. Table~\ref{success} reports task success rates with and without replanning in simulation and the real world. We observe a positive correlation between simulation and real-world success rates. Simulation success rates are slightly higher than in real-world deployment due to the sim-to-real gap. Compared with results without replanning, simulation-based replanning yields higher success rates: object manipulation (cube and ducks) increases by 20 percentage points, while sloth packing increases by 40 percentage points. For real-robot rollouts, the success rate is lower than in replanning-enabled simulation but higher than in both simulation and real-world deployment without replanning. The average success rates across all tasks in simulation and real-world deployment are 66.7\% and 55.3\%, respectively.  With simulation-based replanning, they increase to 93.3\% and 80\%, respectively.

\begin{figure*}[h]
    \vspace{0.5em}
    \centering
    \includegraphics[width=1.0\textwidth]{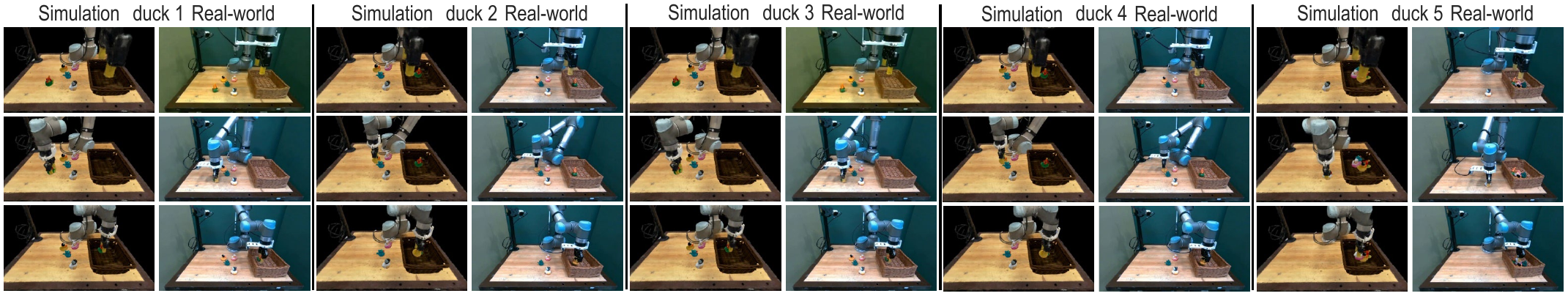}
    \caption{\textbf{Simulation and real-world experiments for duck toy rearrangement}. Left$\rightarrow$Right shows results for ducks 1--5. ``Simulation'' and ``Real-world'' show results from policy simulation and policy rollout on the physical robot.
    } 
    \label{experiment-duck}
 \vspace{-1.0em}
\end{figure*}

Toy duck rearrangement is a long-horizon, multi-stage task in which each episode requires placing five ducks into a basket, with each duck requiring three control steps (e.g., grasp, move, and release). We record the subtask success rate over 10 trials and normalise the results to [0, 1], as shown in Fig.~\ref{success-rate} (right). The results demonstrate that success rates degrade substantially as subtasks and stages progress. The task success rate in simulation is higher than in real-world deployment, reflecting the common sim-to-real gap. Replanning in both simulation and real-world deployment yields higher subtask-level consistency. This improvement is attributable to simulation and replanning, which assess task feasibility prior to action. Fig.~\ref{experiment-duck} shows results for ducks 1--5 in simulation and real-world deployment from left to right. The top, middle, and bottom rows show the initial scene, grasping the duck, and placing the duck into the basket, respectively, in both the simulation and the real world.

\section{Conclusions}
We present Robot-GST, a geometry-aware spatio-temporal behaviour representation and evaluation framework for robotic manipulation that enables simulation-based behaviour verification before execution. Robot-GST reconstructs metrically aligned, photorealistic scenes from short RGB-D captures using 3D Gaussian Splatting with SAM3D-based object understanding, combining photorealistic rendering with explicit geometric proxies for collision and contact reasoning. For language-specified tasks, it grounds long-horizon behaviour generation in 3D keypoint affordances and relational constraints, simulates candidate behaviours, and evaluates them in the reconstructed environment. During execution, Gaussian-aware final-state estimation provides target states for efficient low-level trajectory planning, reducing error accumulation. Experiments on rigid, soft, and deformable tasks demonstrate improved closed-loop reliability and a positive relationship between simulation and real-world performance. Limitations arise from reconstruction and physics fidelity issues under occlusion, dynamic scene changes, and imperfect contact or deformation models, motivating future work on improved physical modelling and uncertainty-aware planning.

\bibliographystyle{IEEEtran}
\bibliography{main}


\end{document}